\documentclass[sigconf]{acmart}
\usepackage{multirow}
\usepackage{tabularx} 
\usepackage{graphicx} 
\usepackage{makecell} 
\usepackage{booktabs}
\usepackage[table]{xcolor}
\usepackage{amsmath}
\usepackage{array}
\AtBeginDocument{%
  }

\setcopyright{acmlicensed}

\copyrightyear{2026}
\acmYear{2026}
\setcopyright{cc}
\setcctype{by}
\acmConference[MM '26]{Proceedings of the 34th ACM International Conference on Multimedia}{November 10--14, 2026}{Rio de Janeiro, Brazil}
\acmBooktitle{Proceedings of the 34th ACM International Conference on Multimedia (MM '26), November 10--14, 2026, Rio de Janeiro, Brazil}
\acmDOI{10.1145/3767308.3836634}
\acmISBN{979-8-4007-2213-4/2026/11}
\begin{document}


\title{Beyond Exact Match: Task-Aware GRPO for Cross-Domain PCBA Visual Question Answering}


\author{Jia Li}
\affiliation{%
  \institution{Hefei University of Technology}
  \city{Hefei}
  \country{China}
}
\email{jiali@hfut.edu.cn}

\author{Li Dai}
\affiliation{%
  \institution{Hefei University of Technology}
  \city{Hefei}
  \country{China}
}
\email{2025110483@mail.hfut.edu.cn}

\author{Peng Jia}
\affiliation{%
  \institution{Hefei University of Technology}
  \city{Hefei}
  \country{China}
}
\email{2020214631@mail.hfut.edu.cn}

\author{Zhenzhen Hu}
\authornote{Corresponding author.}
\affiliation{%
    \institution{Hefei University of Technology}
    \institution{Intelligent Interconnected Systems Laboratory of Anhui Province}
 \city{Hefei}
 \country{China}
}
\email{zzhu@hfut.edu.cn}


\author{Chee Seng Chan}
\affiliation{%
  \institution{Universiti Malaya}
  \city{Kuala Lumpur}
  \country{Malaysia}
}
\email{cs.chan@um.edu.my}

\author{Bingkun Bao}
\affiliation{%
  \institution{Hefei University of Technology}
  \city{Hefei}
  \country{China}
}
\email{bingkunbao@hfut.edu.cn}

\author{Richang Hong}
\affiliation{%
  \institution{Hefei University of Technology}
  \city{Hefei}
  \country{China}
}
\email{hongrc.hfut@gmail.com}

\renewcommand{\shortauthors}{Jia Li et al.}


\begin{abstract}

In automated Printed Circuit Board Assembly (PCBA) inspection, standards-guided decisions require systems to jointly reason over fine-grained visual cues, component semantics, and manufacturing knowledge. Although large vision-language models (VLMs) provide a promising foundation, their deployment is hindered by the domain shift between standards-derived samples and real-world production-line imagery, together with heterogeneous output spaces spanning choice-based and numerical counting tasks. To address these challenges, we propose a multimodal reasoning framework for cross-domain PCBA visual question answering. The framework converts standards-derived, real-world, and auxiliary PCB-domain data into a unified instruction format and constructs verified reasoning traces aligned with visual evidence, question semantics, candidate options, and ground-truth answers. We further introduce Task-Aware Group Relative Policy Optimization(GRPO), which moves beyond exact-match supervision by integrating multi-component semantic rewards for choice-based questions, distance-aware rewards for counting questions, and an auxiliary format reward for valid outputs. During inference, answer--option semantic consistency correction, self-consistency voting, and multi-model arbitration are combined to improve prediction robustness. The proposed system achieves an Overall Score of 83.24 on the official PCBA Standard-to-Real Grand Challenge leaderboard, demonstrating the effectiveness of task-aware reward design and robust inference for cross-domain PCBA visual question answering.
\end{abstract}





\maketitle

\begin{figure}[t]
    \centering
    \includegraphics[width=\linewidth]{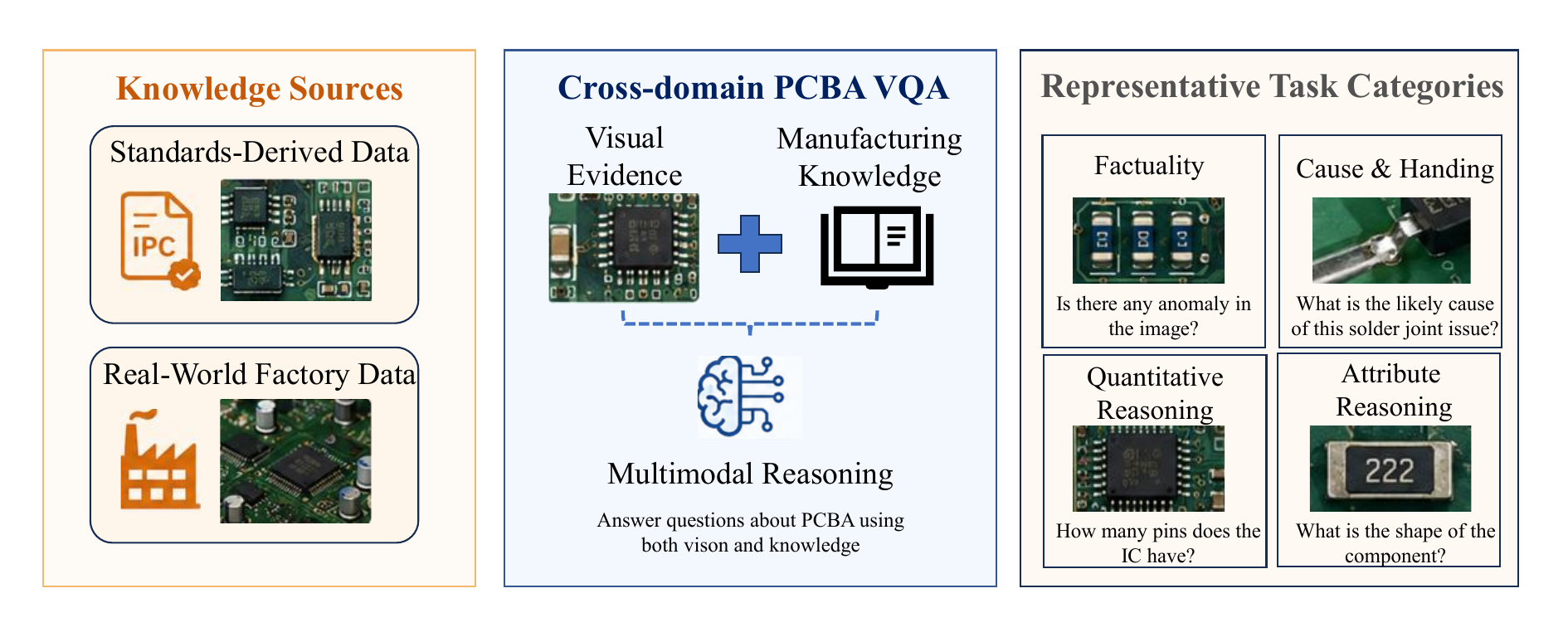}
    \caption{Illustration of cross-domain PCBA visual question answering, which combines standards-derived and real-world factory data and evaluates multimodal reasoning across factuality, cause and handling, quantitative reasoning, and attribute reasoning tasks.}
    \label{introduction——task}
\end{figure}

\section{Introduction}

Industrial visual inspection is essential in Printed Circuit Board Assembly (PCBA), where subtle defects may cause functional failures, reliability degradation, and costly rework~\cite{petkov2024printed}. In Surface Mount Technology (SMT) lines, Automated Optical Inspection (AOI) is increasingly expected to support standards-aligned decisions by connecting fine-grained visual evidence with component semantics, process rules, and handling actions~\cite{goti2025automated,jessurun2023fpic}. Although existing industrial inspection benchmarks and PCB-oriented datasets have advanced anomaly localization, defect detection, and board-level visual understanding~\cite{bergmann2019mvtec,zou2022spot,jiang2410mmad,fan2025manta,tang2019online,lv2024dataset,mineo2025pcb}, most remain perception-oriented and rarely assess whether models can integrate visual observations with manufacturing knowledge to produce actionable decisions. This gap motivates a unified multimodal reasoning framework that combines visual perception, domain knowledge, and structured decision-making.

To move beyond perception-centric inspection, models should be evaluated on their ability to connect visual evidence with manufacturing knowledge in decision-oriented scenarios. The PCBA Standard-to-Real Challenge formulates this problem as cross-domain visual question answering~\cite{chen2026pcba}, combining standards-derived and real-world production-line samples across diverse inspection-oriented question families. This setting requires models to generalize from standards-derived samples to visually diverse production-line images while handling heterogeneous answer spaces and task-specific output constraints. Recent large vision-language models have demonstrated strong multimodal instruction-following and reasoning capabilities~\cite{li2023blip,liu2023visual,bai2025qwen25vltechnicalreport,chen2024expanding}, while Chain-of-Thought(CoT) reasoning and reinforcement-based optimization have shown promising performance in complex reasoning and verifiable problem solving~\cite{wei2022chain,zhang2023multimodal,shao2024deepseekmath,liu2025visual}. Nevertheless, directly applying general-purpose VLMs or generic reasoning optimization pipelines to PCBA VQA remains suboptimal, because valid predictions must be consistent not only with visual evidence, but also with option semantics, numerical targets, and required answer formats.

To address these challenges, we propose a task-aware multimodal framework for PCBA VQA. We unify heterogeneous data sources and question types into a consistent multimodal instruction format and construct verified Chain-of-Thought data aligned with visual evidence, question semantics, candidate options, and final answers. Building on these structured data, we introduce Task-Aware GRPO with answer-type-specific rewards: multi-component semantic rewards for choice-based questions, distance-aware rewards for counting questions, and an auxiliary format reward for valid and parseable responses. During inference, semantic consistency correction, self-consistency voting, and multi-model arbitration are combined to improve robustness. Our method ranks second on the official PCBA Standard-to-Real Challenge leaderboard with an Overall Score of $83.24$.

Our main contributions are summarized as follows:
\begin{itemize}
\item We develop a task-structured data construction pipeline that unifies heterogeneous PCBA samples into a consistent multimodal instruction format and produces verified CoT data.

\item We introduce Task-Aware GRPO with semantic rewards for choice-based reasoning, distance-aware rewards for counting, and format regularization for valid outputs.

\item Our system ranks second on the PCBA Challenge leaderboard with an Overall Score of $83.24$, demonstrating its effectiveness for cross-domain manufacturing inspection VQA.
\end{itemize}

\begin{figure*}[t]
    \centering
    \includegraphics[width=\textwidth]{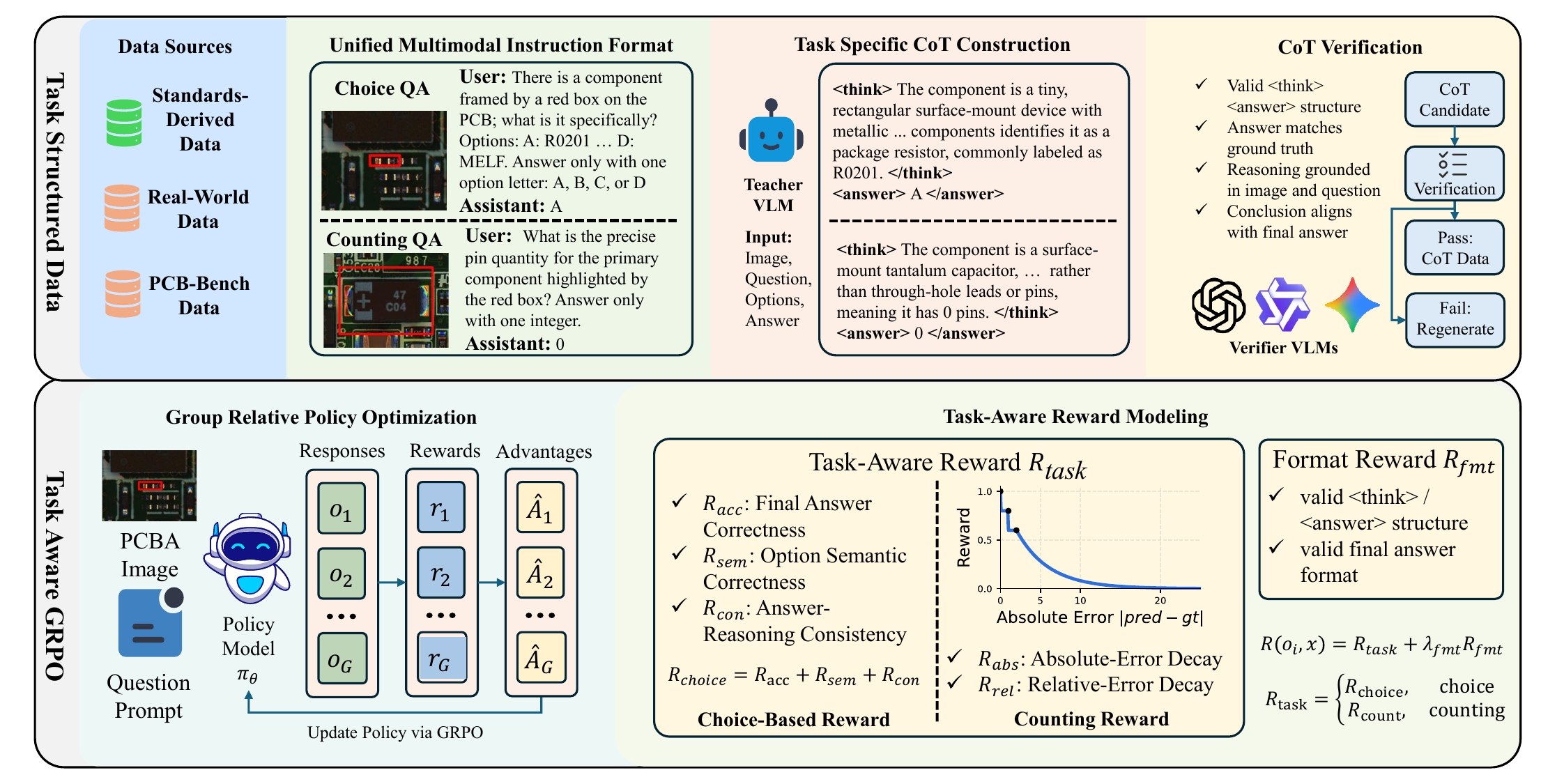}
    \caption{The overall framework consists of two main stages: task-structured data construction and Task-Aware GRPO. In the first stage, standards-derived data, real-world data, and PCB-Bench data are converted into a unified multimodal instruction format, followed by task-specific CoT construction and verification. In the second stage, the policy model is optimized with GRPO using task-aware rewards, including choice-based rewards, counting rewards, and format rewards.}
    \label{fig:model_architecture}
\end{figure*}

\section{Related Work}

\textbf{Industrial Visual Inspection and PCBA Defect Understanding.}
Industrial inspection has been widely studied through anomaly detection and defect localization. Benchmarks such as MVTec AD, MVTec LOCO AD, MVTec AD 2, and Real-IAD extend industrial anomaly detection beyond normal-only defect spotting to settings involving logical anomalies, realistic conditions, and diverse industrial variations~\cite{bergmann2019mvtec,bergmann2022beyond,heckler2503mvtec,wang2024real}. Representative methods have evolved from distribution modeling and self-supervised tasks to reconstruction-, distillation-, and memory-based detectors, including PaDiM, DRAEM, PatchCore, Reverse Distillation, and EfficientAD~\cite{defard2021padim,zavrtanik2021draem,roth2022towards,deng2022anomaly,batzner2024efficientad}. PCB-specific datasets and surveys, such as HRIPCB, the Scientific Data PCB defect dataset, and PCB defect detection reviews, further adapt this paradigm to board-level inspection~\cite{huang2020hripcb,lv2024dataset,chen2023comprehensive}. However, these works mainly predict anomaly existence, location, or category. PCBA instead requires component semantics, defect reasoning, counting, attribute understanding, and standard-guided decisions, motivating a unified multimodal reasoning framework beyond standalone defect detection.

\textbf{Multimodal VQA and Vision-Language Models for Industrial Reasoning.}
Visual question answering has progressed from generic image-question answering to structured and knowledge-intensive multimodal reasoning. VQA, GQA, OK-VQA, TextVQA, and DocVQA establish open-ended visual QA, compositional reasoning, external knowledge, reading, and grounding requirements~\cite{antol2015vqa,hudson2019gqa,marino2019ok,singh2019towards,mathew2021docvqa}. ScienceQA, MathVista, and MMMU further push multimodal models toward scientific, mathematical, and expert-level reasoning~\cite{lu2022learn,lu2024mathvista,yue2024mmmu}. Meanwhile, vision-language models have advanced from CLIP-style contrastive pretraining to BLIP-2, LLaVA, Qwen2.5-VL, and InternVL-style scaling, improving instruction following, OCR, grounding, and high-resolution perception~\cite{radford2021learning,li2023blip,liu2023visual,bai2025qwen25vltechnicalreport,chen2024expanding}. Nevertheless, industrial PCBA inspection differs from generic VQA: PCB-Bench focuses on PCB design and routing reasoning, whereas PCBA Standard-to-Real emphasizes cross-domain generalization from manufacturing standards to real inspection images with mixed answer spaces and tiny visual cues~\cite{li2026pcb,chen2026pcba}. 

\textbf{Chain-of-Thought Reasoning and Reinforcement Optimization.}
Recent reasoning studies show that exposing intermediate steps improves complex problem solving. Chain-of-Thought prompting, zero-shot CoT, STaR, and Tree-of-Thoughts show that explicit reasoning traces help decomposition, planning, and self-correction~\cite{wei2022chain,kojima2022large,zelikman2022star,yao2023tree}. This idea also extends to multimodal reasoning, where Multimodal-CoT and visual CoT prompting align visual evidence with semantic decisions, and step-level verification highlights the value of supervising reasoning processes~\cite{lightman2024let,zhang2023multimodal,chen2024visual}. In parallel, large-model optimization has progressed from RLHF and PPO-based policy learning to direct preference optimization and reasoning-oriented reinforcement learning~\cite{ouyang2022training,schulman2017proximal,rafailov2023direct}. DeepSeekMath introduces GRPO for scalable verifiable reasoning, while DeepSeek-R1 and Visual-RFT show that reinforcement fine-tuning is practical for multimodal reasoning~\cite{shao2024deepseekmath,guo2025deepseek,liu2025visual}. However, most CoT and RL pipelines assume homogeneous targets, whereas PCBA inspection mixes multiple-choice recognition, binary defect judgment, counting, and standard-aware decisions, requiring task-aware reasoning and optimization.

\section{Method}

\subsection{Problem Formulation and Overview}

Given one or more images and a natural-language question, the PCBA task requires a multimodal model to predict an answer. It involves standards-based knowledge, real-world visual recognition, attribute reasoning, and quantitative reasoning, requiring stable reasoning in complex scenarios. Based on their output formats, we categorize questions into three types: multiple-choice, binary judgment, and numerical counting. Choice and binary questions correspond to multi-class and two-class judgments and require selecting a valid option from visual evidence and question semantics; counting questions identify target objects and return an integer.

Our framework consists of three components. First, to reduce format shift across heterogeneous sources and question types, we organize all samples into a unified multimodal instruction format and construct verified CoT-format reasoning data, providing a consistent training basis for supervised fine-tuning and reasoning optimization. Second, to address heterogeneous answer spaces in PCBA VQA, we introduce Task-Aware GRPO with answer-type-specific reward functions. It combines hierarchical rewards for choice-based reasoning, distance-aware rewards for counting reasoning, and format compliance rewards, enabling reinforcement optimization to jointly improve answer correctness, semantic consistency, numerical accuracy, and instruction-following ability. Third, during inference, we combine semantic consistency correction, self-consistency estimation, and multi-model arbitration to improve robustness.

\subsection{Task Structured Data Construction}

\textbf{Unified Multimodal Instruction Format.}
PCBA VQA contains heterogeneous question types that differ in question formulations and output constraints. Directly mixing these samples may introduce format shift and make the model overly sensitive to superficial prompt patterns. To mitigate this issue, we convert samples from different sources and task categories into a unified multimodal instruction format. For choice-based questions, the output is normalized to a single valid option letter; for counting questions, the output is normalized to an integer without additional explanations or redundant units. This design has two benefits. First, it provides a consistent training interface for standards-derived knowledge understanding, defect judgment, and quantitative reasoning. Second, it constrains the output space, thereby improving instruction following and providing a stable basis for reward design.

\textbf{Task Specific CoT Construction and Verification.}
To enhance intermediate reasoning on complex PCBA questions, we construct CoT-format reasoning data for subsequent reasoning-oriented optimization. Each CoT sample consists of a concise reasoning process and a final answer. The reasoning process is expected to focus on visual evidence, question semantics, or standards-related cues directly relevant to the current sample, while the final answer follows the required answer format.

During CoT construction, we use a strong vision-language model and provide it with the image, the question, candidate options when available, and the ground-truth answer. This setting helps keep the generated reasoning process consistent with the final answer and reduces hallucinated explanations. For different answer spaces, we adopt task-specific generation instructions: choice-based samples emphasize option-semantic alignment, while counting samples emphasize numerical consistency. Furthermore, we introduce a verification step to improve the reliability of CoT data. The verification process checks whether the output follows the required CoT structure, whether the final answer matches the ground-truth answer, whether the reasoning content is grounded in the visual evidence and question semantics, and whether the reasoning conclusion is consistent with the candidate option or integer answer. Samples that fail verification are regenerated or filtered out. Through this process, we obtain reliable and parseable reasoning data for subsequent Task-Aware GRPO optimization.

\subsection{Task-Aware GRPO}

To further improve performance on PCBA VQA tasks, we apply Group Relative Policy Optimization (GRPO)~\cite{shao2024deepseekmath}. Given an input sample $x$, the old policy $\pi_{\theta_{\mathrm{old}}}$ samples a group of responses $\{o_i\}_{i=1}^{G}$. Each response is evaluated by a reward function, yielding a scalar reward $r_i=R(o_i,x)$. The relative advantage of the $i$-th response is computed by normalizing rewards within the sampled group:
\begin{equation}
\hat{A}_i =
\frac{
r_i-\operatorname{mean}(\{r_j\}_{j=1}^{G})
}{
\operatorname{std}(\{r_j\}_{j=1}^{G})+\epsilon
},
\end{equation}
where $\epsilon$ is a small constant for numerical stability. This group-relative normalization removes the need for a separate value model and encourages the policy to favor responses that receive higher rewards than other responses sampled for the same input.

The policy is optimized with a clipped objective:
\begin{equation}
\mathcal{J}_{\mathrm{GRPO}}(\theta)
=
\mathbb{E}_{x,\{o_i\}_{i=1}^{G}}
\left[
\frac{1}{G}
\sum_{i=1}^{G}
S_i
-
\beta
D_{\mathrm{KL}}
\left(
\pi_{\theta}(\cdot|x)
\|
\pi_{\mathrm{ref}}(\cdot|x)
\right)
\right],
\end{equation}
where
\begin{equation}
S_i =
\min
\left(
\rho_i \hat{A}_i,
\operatorname{clip}(\rho_i,1-\epsilon_c,1+\epsilon_c)\hat{A}_i
\right),
\end{equation}
and
\begin{equation}
\rho_i =
\frac{\pi_{\theta}(o_i|x)}
{\pi_{\theta_{\mathrm{old}}}(o_i|x)}.
\end{equation}
Here, $\rho_i$ denotes the policy ratio, $\epsilon_c$ is the clipping threshold, $\pi_{\mathrm{ref}}$ denotes the reference policy, and $\beta$ controls the strength of KL regularization.


\subsubsection{Task-Aware Reward Modeling}

Since PCBA VQA contains heterogeneous answer spaces, a single generic reward is insufficient for optimizing different question types with distinct evaluation criteria. We therefore design task-aware rewards for choice-based and counting tasks, together with a format reward to encourage valid and instruction-following outputs. For each sampled response $o_i$, the overall reward is defined as:
\begin{equation}
R(o_i,x)
=
R_{\mathrm{task}}(o_i,x)
+
\lambda_{\mathrm{fmt}}R_{\mathrm{fmt}}(o_i,x),
\end{equation}
where $R_{\mathrm{task}}$ denotes the task-specific reward selected according to the question type, $R_{\mathrm{fmt}}$ denotes the format reward, and $\lambda_{\mathrm{fmt}}$ controls the strength of the format constraint. $R_{\mathrm{task}}$ is defined as:
\begin{equation}
R_{\mathrm{task}}(o_i,x)
=
\begin{cases}
R_{\mathrm{choice}}(o_i,x), & \text{choice-based tasks},\\
R_{\mathrm{count}}(o_i,x), & \text{counting tasks}.
\end{cases}
\end{equation}

\paragraph{Choice-Based Reward.}
Choice-based tasks include both multiple-choice questions and binary judgment questions. Let $\hat{a}_i$ and $a$ denote the predicted and ground-truth option labels for response $o_i$, respectively. Let $\hat{e}_i$ denote the semantic conclusion extracted from the reasoning trace, while $e_a$ and $e_{\hat{a}_i}$ denote the option texts corresponding to the ground-truth label and the predicted label. We define three reward components:
\begin{equation}
\left\{
\begin{array}{ll}
R_{\mathrm{acc}}(o_i,x)=1, & \text{if } \hat{a}_i=a, \text{ otherwise } 0,\\
R_{\mathrm{sem}}(o_i,x)=1, & \text{if } \hat{e}_i\simeq e_a, \text{ otherwise } 0,\\
R_{\mathrm{con}}(o_i,x)=1, & \text{if } \hat{e}_i\simeq e_{\hat{a}_i}, \text{ otherwise } 0.
\end{array}
\right.
\end{equation}
where $\simeq$ denotes semantic equivalence. Here, $R_{\mathrm{acc}}$ measures final answer correctness, $R_{\mathrm{sem}}$ measures semantic alignment with the ground-truth option text, and $R_{\mathrm{con}}$ measures consistency between the predicted option and the reasoning conclusion.

The final choice-based reward is computed as:
\begin{equation}
R_{\mathrm{choice}}(o_i,x)
=
\alpha_{\mathrm{acc}}R_{\mathrm{acc}}
+
\alpha_{\mathrm{sem}}R_{\mathrm{sem}}
+
\alpha_{\mathrm{con}}R_{\mathrm{con}},
\end{equation}
where $\alpha_{\mathrm{acc}}$, $\alpha_{\mathrm{sem}}$, and $\alpha_{\mathrm{con}}$ are weighting coefficients. This design encourages the model to align option labels with their semantic meanings rather than relying only on superficial label patterns.

\paragraph{Counting Reward.}
For counting tasks, exact-match rewards are overly sparse, since near-correct numerical predictions should be distinguished from large counting errors. Let $\hat{y}_i$ and $y$ denote the predicted and ground-truth counts for response $o_i$, respectively. We define the absolute error and, when $y\neq0$, the relative error as:
\begin{equation}
d_i = |\hat{y}_i-y|,
\qquad
\delta_i = \frac{d_i}{|y|}\quad (y\neq0).
\end{equation}

To provide smoother supervision, we assign fixed rewards to exact and near-correct predictions, while applying error-dependent decay to larger deviations. Specifically, we define absolute-error and relative-error decay terms as:
\begin{equation}
R_{\mathrm{abs}}(d_i,y)
=
0.6
\exp
\left(
-\frac{d_i-2}{s_{\mathrm{abs}}(y)}
\right),
\quad
R_{\mathrm{rel}}(\delta_i)
=
0.6
\left(
1-\frac{\delta_i}{\tau_{\mathrm{rel}}}
\right)_+^2,
\end{equation}
where $(z)_+=\max(z,0)$, $s_{\mathrm{abs}}(y)$ controls the absolute-error decay scale, and $\tau_{\mathrm{rel}}$ denotes the relative-error tolerance. The counting reward is then defined as:
\begin{equation}
R_{\mathrm{count}}(o_i,x)
=
\begin{cases}
1.0, & d_i=0,\\
0.8, & d_i=1,\\
0.6, & d_i=2,\\
\min\left(R_{\mathrm{abs}}(d_i,y),R_{\mathrm{rel}}(\delta_i)\right), & d_i>2,\ y\neq0,\\
R_{\mathrm{abs}}(d_i,0), & d_i>2,\ y=0.
\end{cases}
\end{equation}
When $y=0$, only the absolute-error decay is used to avoid undefined relative error. Compared with an exact-match reward alone, this design provides denser supervision for near-correct counts and penalizes large absolute and relative errors. By producing more informative reward differences within sampled response groups, it yields more stable relative advantages for GRPO training.

\paragraph{Format Reward.}
To preserve output validity, we introduce a format reward $R_{\mathrm{fmt}}$. This reward checks two aspects: whether the final answer conforms to the answer space specified by the input $x$, and whether the response follows a complete reasoning-answer structure. Let $v_a(o_i,x)\in\{0,1\}$ indicate whether the answer in response $o_i$ is valid, and let $v_s(o_i)\in\{0,1\}$ indicate whether the reasoning-answer structure is complete. The format reward is defined as:
\begin{equation}
R_{\mathrm{fmt}}(o_i,x) =
\begin{cases}
1.0, & v_a(o_i,x)=1 \ \text{and}\ v_s(o_i)=1,\\
0.5, & v_a(o_i,x)=1 \ \text{and}\ v_s(o_i)=0,\\
0.2, & v_a(o_i,x)=0 \ \text{and}\ v_s(o_i)=1,\\
0, & v_a(o_i,x)=0 \ \text{and}\ v_s(o_i)=0.
\end{cases}
\end{equation}
This auxiliary reward encourages instruction-following responses. Moreover, its graded piecewise design provides non-binary feedback for partially valid outputs, which reduces reward sparsity and improves the stability of reinforcement learning optimization.

\subsection{Confidence-Calibrated Robust Inference}

During inference, we apply a lightweight calibration strategy to reduce answer inconsistency and sampling instability. For choice-based tasks, if the reasoning conclusion is semantically aligned with one candidate option but the answer corresponds to another option, we correct the answer to the semantically matched option.

We further estimate confidence through self-consistency voting. Given $K$ sampled answers $\{\hat{z}^{(k)}\}_{k=1}^{K}$, the majority-vote answer and its confidence are defined as:
\begin{equation}
\begin{aligned}
\hat{z}^{*}
&=
\operatorname*{arg\,max}_{v}
\left|\{k \mid \hat{z}^{(k)}=v,\ 1\leq k\leq K\}\right|,\\
c_{\mathrm{vote}}
&=
\frac{1}{K}
\max_{v}
\left|\{k \mid \hat{z}^{(k)}=v,\ 1\leq k\leq K\}\right|.
\end{aligned}
\end{equation}
If $c_{\mathrm{vote}}$ exceeds a confidence threshold, we adopt $\hat{z}^{*}$ as the answer. Otherwise, the sample is treated as low-confidence and further resolved by multi-model arbitration, where predictions from multiple vision-language models are aggregated by majority voting. This strategy improves inference robustness by combining semantic consistency, sampling consistency, and model-level agreement.

\subsection{Training Details}

We use Qwen3.5-27B~\cite{qwen35blog} as the backbone model and adopt LoRA~\cite{hu2022lora} for parameter-efficient fine-tuning. Training is conducted in two stages. In the first stage, supervised fine-tuning is performed on unified multimodal instruction data to align the model with PCB-domain knowledge and task formats. In the second stage, Task-Aware GRPO is applied to task-structured CoT data to further improve reasoning performance across heterogeneous PCBA VQA tasks, including choice-based and counting questions.

\begin{table*}[t]
\centering
\caption{Comparison with the top five teams on the final PCBA Standard-to-Real Grand Challenge leaderboard. Our method is highlighted in gray and boldface. All scores are reported as percentages.}
\label{tab:competition_results}
\resizebox{\textwidth}{!}{
\begin{tabular}{c l c c c c c c c c c}
\toprule
Rank & Submission & Overall & Attribute & Comp. Type & Count Comp. & Count Pin & Defect Exist. & Defect Type & Mount Side & Standard Know. \\
\midrule
1 & syds3sxhdjs & 83.82 & 100.00 & 95.08 & 98.96 & 99.55 & 81.25 & 85.25 & 87.50 & 66.00 \\
\rowcolor{gray!15}
2 & \textbf{Ours} & \textbf{83.24} & \textbf{100.00} & \textbf{90.16} & \textbf{98.87} & \textbf{98.11} & \textbf{82.47} & \textbf{84.43} & \textbf{87.50} & \textbf{66.00} \\
3 & travis111 & 81.33 & 100.00 & 95.08 & 98.99 & 99.52 & 80.85 & 79.51 & 91.67 & 60.00 \\
4 & gmehong & 80.11 & 100.00 & 96.72 & 98.96 & 98.13 & 76.92 & 77.05 & 91.67 & 61.00 \\
5 & dkhonker & 77.52 & 100.00 & 93.44 & 98.72 & 97.58 & 77.08 & 73.77 & 75.00 & 58.00 \\
\bottomrule
\end{tabular}
}
\end{table*}

\begin{table*}[t]
\centering
\caption{Ablation studies on the public evaluation set, evaluating backbone scale, auxiliary data, and GRPO reward design. All scores are reported as percentages, and the best result within each ablation group is shown in boldface.}
\label{tab:ablation_studies}
\small
\setlength{\tabcolsep}{4pt}
\resizebox{\textwidth}{!}{
\begin{tabular}{l c c c c c c c c c}
\toprule
Setting & Overall & Attribute & Comp. Type & Count Comp. & Count Pin & Defect Exist. & Defect Type & Mount Side & Standard Know. \\
\midrule
\multicolumn{10}{l}{\textbf{Backbone Scale and Auxiliary Data}} \\
Qwen3.5-4B 
& 82.66 & 80.25 & 73.80 & 97.81 & 98.45 & 94.82 & 81.15 & \textbf{87.25} & 66.63 \\
Qwen3.5-9B 
& 83.99 & 98.75 & 83.90 & 97.84 & 97.66 & 93.91 & 77.75 & 84.50 & 68.75 \\
Qwen3.5-27B 
& 86.84 & \textbf{100.00} & 89.00 & \textbf{98.35} & 98.69 & 94.78 & \textbf{82.50} & 84.75 & 72.38 \\
Qwen3.5-27B + PCB-Bench 
& \textbf{86.93} & 99.50 & \textbf{90.10} & 97.55 & \textbf{98.89} & \textbf{95.09} & 82.05 & 84.50 & \textbf{72.69} \\
\midrule
\multicolumn{10}{l}{\textbf{GRPO Reward Design}} \\
SFT 
& 86.93 & \textbf{99.50} & 90.10 & 97.55 & 98.89 & 95.09 & 82.05 & 84.50 & 72.69 \\
GRPO w/ format reward 
& 87.01 & 96.75 & 90.10 & 98.02 & 98.19 & 94.91 & 82.65 & 84.75 & 73.25 \\
GRPO w/ task-aware reward 
& 88.72 & 97.50 & \textbf{92.60} & 98.35 & 98.89 & \textbf{95.71} & 86.15 & 88.75 & \textbf{73.63} \\
\textbf{Ours} 
& \textbf{88.75} & 97.50 & 92.50 & \textbf{98.37} & \textbf{98.92} & 95.62 & \textbf{86.35} & \textbf{89.00} & \textbf{73.63} \\
\bottomrule
\end{tabular}
}
\end{table*}

\section{Experiments}

\subsection{Dataset and Evaluation Protocol}

We use the PCBA SMT Dataset from the PCBA Challenge~\cite{chen2026pcba} and incorporate PCB-Bench~\cite{li2026pcb} as auxiliary data. The PCBA SMT Dataset covers four task categories: Defect Cause \& Handling, Perception-Level Factuality, Quantitative Reasoning, and Attribute Reasoning. The training set contains $8{,}200$ samples, including $1{,}600$ standards-derived samples and $6{,}600$ real-world samples, while the public test set contains $8{,}200$ samples. We additionally use $2{,}193$ samples from PCB-Bench. After unified format conversion, the SFT and CoT training set contains $10{,}393$ samples, comprising $9{,}518$ choice-based questions and $875$ counting questions. For CoT construction, Qwen3.6~\cite{qwen3.6-27b} serves as the teacher VLM, while GPT-5.4~\cite{openai2026gpt54}, Qwen3.6~\cite{qwen3.6-27b}, and Gemini~\cite{geminiteam2023gemini} are used as verifier VLMs.

We follow the evaluation protocol of the PCBA Challenge~\cite{chen2026pcba}. Choice-based tasks are evaluated using accuracy. For defect-existence questions, accuracy and F1-score are averaged to account for potential class imbalance. Counting tasks are evaluated using exact accuracy and mean absolute error (MAE), where MAE is converted into a normalized higher-is-better score and averaged with accuracy. The Overall Score is computed as the unweighted mean of all question-family scores. Let $S_{\mathrm{test}}$ and $S_{\mathrm{hidden}}$ denote the aggregated scores on the public test split and hidden split, respectively, and the final score is defined as $S_{\mathrm{final}} = 0.8S_{\mathrm{test}} + 0.2S_{\mathrm{hidden}}$. Both the public test set and hidden test set are used only for evaluation.

\subsection{Experimental Setup}

All experiments are conducted on $4$ H100 GPUs. Training follows a two-stage pipeline consisting of supervised fine-tuning and GRPO-based reasoning optimization. In the GRPO stage, the format reward weight is set to $\lambda_{\mathrm{fmt}}=0.1$. For choice-based rewards, the component weights are set to $\alpha_{\mathrm{acc}}=0.70$, $\alpha_{\mathrm{sem}}=0.20$, and $\alpha_{\mathrm{con}}=0.10$.

In the SFT stage, we set the LoRA rank to $16$, LoRA alpha to $32$, the learning rate to $5\times10^{-5}$, and train the model for $2$ epochs. In the GRPO stage, we set the LoRA rank to $8$, LoRA alpha to $32$, the learning rate to $5\times10^{-7}$, the rollout group size to $G=8$, the sampling temperature to $1.0$, and train the model for $1$ epoch. During inference, we first perform one greedy decoding pass and then generate $K=8$ stochastic samples for self-consistency voting. The sampling configuration uses temperature $=0.5$ and top-$p=0.9$, with the voting confidence threshold set to $0.55$.

\subsection{Comparison with Competitors}

We compare our method with the top-performing teams on the PCBA Standard-to-Real Challenge test set. As shown in Table~\ref{tab:competition_results}, our method ranks second overall with an Overall Score of $83.24$, only $0.58$ points behind the first-place solution. Notably, our method achieves the best performance on Defect Existence with a score of $82.47$, suggesting that the proposed task-aware optimization and confidence-calibrated inference strategy improve the robustness of defect-existence recognition. We also obtain perfect performance on Attribute Reasoning with a score of $100.00$, competitive results on Component Counting and Pin Counting with scores of $98.87$ and $98.11$, respectively, and a tied-best score of $66.00$ on Standard Knowledge. These results show that our method maintains strong robustness across diverse PCBA VQA tasks, including perception-level factuality, quantitative reasoning, and standards-derived knowledge reasoning. Overall, the second-place result validates the effectiveness of our task-structured data construction, task-aware reward design, and confidence-calibrated inference strategy.

\subsection{Ablation Study}

\paragraph{Backbone Scale and Auxiliary Data.}
We first evaluate the contributions of backbone capacity and auxiliary PCB-domain data. As shown in Table~\ref{tab:ablation_studies}, scaling the backbone from 4B to 27B consistently improves the Overall Score from $82.66$ to $86.84$. The improvement is particularly evident on Component Type, Attribute Reasoning, and Standard Knowledge, suggesting that larger vision-language backbones provide stronger fine-grained visual recognition and domain knowledge modeling capabilities. The 27B model also improves Defect Type recognition, which requires distinguishing subtle local patterns under dense PCB layouts. We further introduce PCB-Bench as auxiliary PCB-domain data. Compared with 27B trained with the base data setting, adding PCB-Bench slightly improves the Overall Score from $86.84$ to $86.93$. The gains are concentrated in Component Type, Defect Existence, Pin Counting, and Standard Knowledge. This indicates that auxiliary PCB-domain data improves domain coverage, while the official PCBA training set remains the primary source of task-specific supervision.

\paragraph{GRPO Reward Design.}
We then ablate the reward design used in the GRPO stage, using the SFT model as the baseline. As reported in Table~\ref{tab:ablation_studies}, applying GRPO using only a simple format reward marginally improves the Overall Score from $86.93$ to $87.01$, indicating that sparse exact-match feedback provides limited supervision for heterogeneous PCBA VQA tasks. In contrast, replacing the format reward with the proposed task-aware reward increases the Overall Score to $88.72$, with clear gains on Component Type, Defect Type, and Mount Side. This suggests that modeling different answer spaces with task-specific rewards is beneficial for both fine-grained recognition and reasoning-oriented questions. Adding the format reward further improves the Overall Score to $88.75$. Although the improvement over the task-aware reward alone is modest, the full reward design achieves the best performance on counting-related metrics, Defect Type, and Mount Side, showing that format regularization helps maintain stable and parseable outputs during reinforcement optimization.

\section{Conclusion}

This study shows that cross-domain PCBA visual question answering is challenged not only by the visual gap between standards-derived and real-world production data, but also by the mismatch between generic exact-match objectives and heterogeneous answer spaces. By combining task-structured data, verified CoT supervision, and Task-Aware GRPO, our framework provides more informative optimization signals for both choice-based and counting questions. The ablation results indicate that answer-type-specific reward modeling contributes most of the performance improvement, while format regularization provides a smaller but consistent gain by maintaining valid and parseable outputs. Together with an Overall Score of 83.24 on the PCBA Standard-to-Real Grand Challenge, these findings support the effectiveness of moving beyond exact-match supervision in industrial multimodal reasoning. More broadly, our results suggest that PCBA inspection-oriented VQA systems benefit when their optimization objectives reflect the semantic and numerical structures of the underlying decisions rather than treating all outputs as homogeneous labels. This finding further emphasizes the importance of task-aware optimization for reliable industrial multimodal reasoning systems.

\begin{acks}
This work was supported in part by National Natural Science Foundation of China under Grant No.U23B2031, and in part by the Fundamental Research Funds for the Central Universities  under Grant No. PA2025IISL0110. The computations were performed on the High-Performance Computing (HPC) Platform of Hefei University of Technology.
\end{acks}

\bibliographystyle{ACM-Reference-Format}
\bibliography{sample-base}

\end{document}